\documentclass[11pt,a4paper]{article}
\usepackage[hyperref]{acl2018}
\usepackage{times}
\usepackage{latexsym}
\usepackage{booktabs}
\usepackage{url}
\usepackage{bm}
\usepackage{bbm}
\usepackage{graphicx}
\usepackage[framemethod=TikZ]{mdframed}
\usepackage{amsmath}
\usepackage{caption}
\usepackage{subcaption}
\usepackage{amsfonts}
\usepackage{amssymb}
\usepackage{breqn}
\usepackage{tikz, tikz-qtree}

\aclfinalcopy % Uncomment this line for the final submission
\def\aclpaperid{1106} %  Enter the acl Paper ID here

\title{Extending a Parser to Distant Domains Using a Few Dozen Partially Annotated Examples}

\author{Vidur Joshi, Matthew Peters, Mark Hopkins \\
  Allen Institute for AI, Seattle, WA \\
  {\tt \{vidurj, matthewp, markh\}@allenai.org}\\}

\date{}

\hypersetup{draft}
\begin{document}
\maketitle

\begin{abstract}
We revisit domain adaptation for parsers in the neural era. First we show that recent advances in word representations greatly diminish the need for domain adaptation when the target domain is syntactically similar to the source domain. As evidence, we train a parser on the Wall Street Journal alone that achieves over 90\% $F_1$ on the Brown corpus. For more syntactically distant domains, we provide a simple way to adapt a parser using only dozens of partial annotations. For instance, we increase the percentage of error-free geometry-domain parses in a held-out set from 45\% to 73\% using approximately five dozen training examples. In the process, we demonstrate a new state-of-the-art single model result  on the Wall Street Journal test set of 94.3\%. This is an absolute increase of 1.7\% over the previous state-of-the-art of 92.6\%. 
\end{abstract}

%\begin{abstract}
%We revisit parser domain adaptation in light of recent developments in neural natural language processing, observing that (a) it is easy to create training data for parser models like \cite{DBLP:journals/corr/SternAK17} that classify spans independently, and (b) the use of contextualized word representations \cite{Peters2017SemisupervisedST} greatly reduces the amount of data needed to train linguistic models. Taking advantage of these developments, we show how to achieve dramatic improvements in domain performance with only dozens of partially annotated sentences, e.g. increasing the percentage of error-free geometry-domain parses in a held-out set from 42\% to 68\% using approximately five dozen training examples. Moreover, the domain-extended parser does not suffer any degradation on newswire parsing, achieving an $F_1$-score on the Wall Street Journal test set of 94.2\%. This is a new state-of-the-art single model result for newswire, an absolute increase of 1.7\% over the previous state-of-the-art of 92.5\%.
%\end{abstract}

\section{Introduction\label{sec:intro}}

Statistical parsers are often criticized for their performance outside of the domain they were trained on. The most straightforward remedy would be more training data in the target domain, but building treebanks \citep{marcus1993building} is expensive.

\begin{figure}[tb]
\begin{mdframed}[roundcorner=10pt]
\textbf{Given [ the circle [ at the right ] with [ designated center, designated perpendicular, and radius 5 ] ] .} \\ \vspace{2mm} \\ \textbf{In [ the figure above ] , [ [ AD = 4 ] , [ AB = 3 ] and [ CD = 9 ] ] .} \\ \vspace{2mm} \\ \textbf{[ Diameter AC ] is perpendicular [ to chord BD ] [ at E ] .}
\end{mdframed}
\caption{\label{fig:markups} An example of partial annotations. Annotators indicate that a span is a constituent by enclosing it in square brackets.}
\end{figure}

In this paper, we revisit this issue in light of recent developments in neural natural language processing. Our paper rests on two observations:
% Our foundation will be the \emph{minimal span parser} \cite{DBLP:journals/corr/SternAK17}, henceforth MSP, which demonstrated state-of-the-art newswire parsing results using two neural classifiers: one that predicts whether a span is a constituent, and one that predicts the label of the span.

\begin{enumerate}
	\item \textbf{It is trivial to train on partial annotations using a span-focused model.} \citet{DBLP:journals/corr/SternAK17} demonstrated that a parser with minimal dependence between the decisions that produce a parse can achieve state-of-the-art performance. We modify their parser, henceforth MSP, so that it trains directly on individual labeled spans instead of parse trees. This results in a parser that can be trained, with no adjustments to the training regime, from partial sentence bracketings.
	\item \textbf{The use of contextualized word representations \citep{Peters2017SemisupervisedST,McCann2017LearnedIT} greatly reduces the amount of data needed to train linguistic models.} Contextualized word representations, which encode tokens conditioned on their context in a sentence, have been shown to give significant boosts across a variety of NLP tasks, and also to reduce the amount of data needed by an order of magnitude in some tasks.
\end{enumerate}

\noindent Taken together, this suggests a way to rapidly extend a newswire-trained parser to new domains. Specifically, we will show it is possible to achieve large out-of-domain performance improvements using only dozens of partially annotated sentences, like those shown in Figure~\ref{fig:markups}. The resulting parser also does not suffer any degradation on the newswire domain.

Along the way, we provide several other notable contributions:

\begin{itemize}
	\item We raise the state-of-the-art single-model $F_1$-score for constituency parsing from 92.6\% to 94.3\% on the Wall Street Journal (WSJ) test set. A trained model is publicly available.\footnote{http://allennlp.org/models}
	\item We show that, even without domain-specific training data, our parser has much less out-of-domain degradation than previous parsers on ``newswire-adjacent" domains like the Brown corpus.
    \item We provide a version of MSP which predicts its own POS tags (rather than requiring a third-party tagger).
\end{itemize}

\section{The Reconciled Span Parser (RSP)}

When we allow annotators to selectively annotate important phenomena, we make the process faster and simpler \citep{Mielens2015ParseIF}. Unfortunately, this produces a disconnect between the model (which typically asserts the probability of a full parse tree) and the annotation task (which asserts the correctness of some subcomponent, like a constituent span or a dependency arc). There is a body of research \citep{Hwa1999SupervisedGI,Li2016TrainingDP} that discusses how to bridge this gap by modifying the training data, training algorithm, or the training objective.

Alternatively, we could just better align the model with the annotation task. Specifically, we could train a parser whose base model predicts exactly what we ask the annotator to annotate, e.g. whether a particular span is a constituent. This makes it trivial to train with partial or full annotations, because the training data reduces to a collection of span labels in either case.

Luckily, recent state-of-the-art results that model NLP tasks as independently classified spans \citep{DBLP:journals/corr/SternAK17} suggest this strategy is currently viable. In this section, we present the Reconciled Span Parser (RSP), a modified version of the Minimal Span Parser (MSP) of \citet{DBLP:journals/corr/SternAK17}. RSP differs from MSP in the following ways:

\begin{itemize}
\item \textbf{It is trained on a span classification task. } 
MSP trains on a maximum margin objective; that is, the loss function penalizes the violation of a margin between the scores of the gold parse and the next highest scoring parse decoded. This couples its training procedure with its decoding procedure, resulting in two versions, a top-down parser and a chart parser. To allow our model to be trained on partial annotations, we change the training task to be the span classification task described below.

\item \textbf{It uses contextualized word representations instead of predicted part-of-speech tags. } Our model uses contextualized word representations as described in \citet{2018arXiv180205365P}. It does not take part-of-speech-tags as input, eliminating the dependence of the parser on a newswire-trained POS-tagger.  
%\item \textbf{We train one model instead of two. } We find no empirical advantage to training separate models for predicting span constituency and span labels. For convenience of domain retraining, we collapse these into a single model. 
% Our model is trained to classify individual spans using a maximum likelihood objective, whereas MSP is trained on a maximum margin objective that depends on having an entire parse tree.
% Developed in the context of chart parsing, MSP classifies the constituent spans of binarized parse trees.  Instead, we align our base model with the more intuitive annotation task by not binarizing our training data. We formulate an Integer Linear Program (ILP) to reconcile the independently chosen constituent spans.

\end{itemize}

\begin{figure*}
\centering
\begin{subfigure}{0.5\textwidth}
  \centering
  \includegraphics[width=\linewidth]{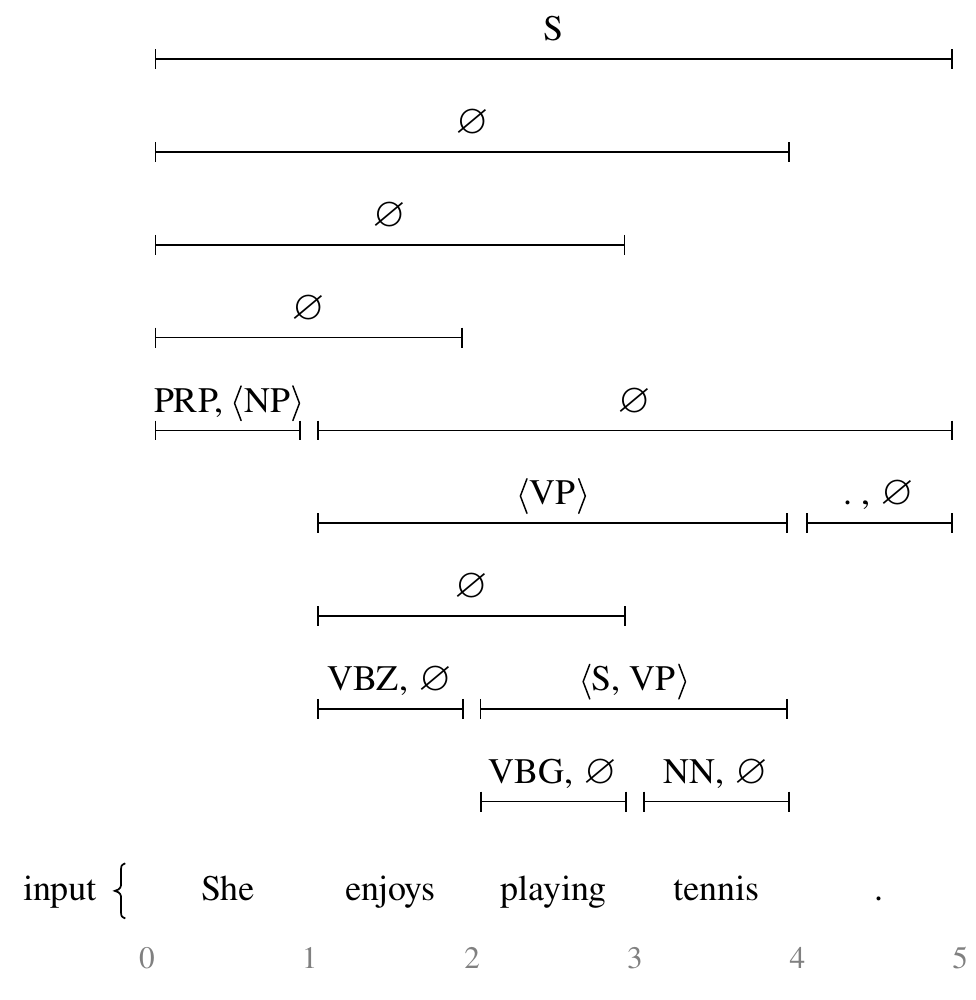}
  \caption{Spans classified by the parsing procedure. Note that leaves have their part-of-speech tags predicted in addition to their sequence of non-terminals.}
  \label{fig:classified-spans}
\end{subfigure} 
\begin{subfigure}{0.49\textwidth}
  \centering
  \includegraphics{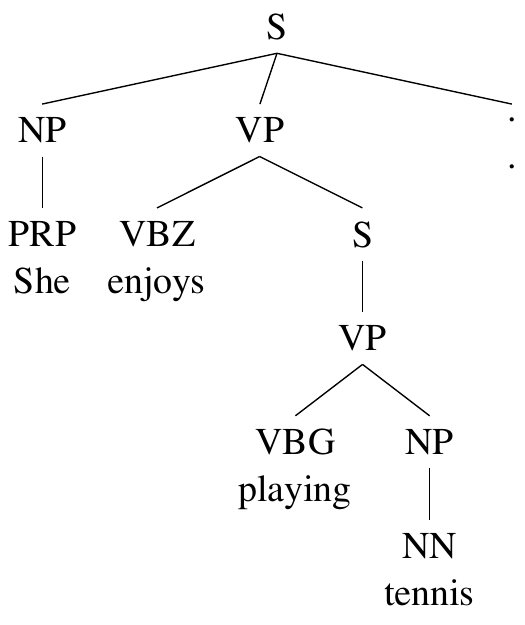}
  \caption{The resulting parse tree.}
  \label{fig:parse-tree}
\end{subfigure}
\caption{The correspondence between labeled spans and a parse tree. This diagram is adapted from figure 1 in \citep{DBLP:journals/corr/SternAK17}.}
\label{fig:top-down-parser}
\end{figure*}

\subsection{Overview}

We will view a parse tree as a labeling of all the spans of a sentence such that:
\begin{itemize}
\item Every constituent span is labeled with the sequence of non-terminals assigned to it in the parse tree. For instance, span $(2, 4)$ in Figure~\ref{fig:parse-tree} is labeled with the sequence $\langle S, \text{VP} \rangle$, as shown in Figure~\ref{fig:classified-spans}. 
\item Every non-constituent is labeled with the empty sequence. 
\end{itemize}

\noindent Given a sentence represented by a sequence of tokens $x$ of length $n$, define $\mathsf{spans}(x) = \{ (i,j) \mid 0 \leq i < j \leq n \}$. Define a \emph{parse} for sentence $x$ as a function $\pi: \mathsf{spans}(x) \mapsto \mathcal{L}$ where $\mathcal{L}$ is the set of all sequences of non-terminal tags, including the empty sequence.

We model the probability of a parse as the independent product of its span labels:

\begin{eqnarray*}
Pr(\pi | x) &=& \prod_{s \in \mathsf{spans}(x)} Pr(\pi(s) \mid x, s) \\
\Rightarrow \log Pr(\pi | x) &=& \sum_{s \in \mathsf{spans}(x)} \log Pr(\pi(s) \mid x, s) 
\end{eqnarray*}

\noindent Hence, we will train a base model $\sigma(l \mid x, s)$ to estimate the log probability of label $l$ for span $s$ (given sentence $x$), and we will score the overall parse with:

\begin{equation*}
\mathsf{score}(\pi | x) = \sum_{s \in \mathsf{spans}(x)} \sigma(\pi(s) \mid x, s)
\end{equation*}

\noindent Note that this probability model accords mass to mis-structured trees (e.g. overlapping spans like (2, 5) and (3, 7) cannot both be constituents of a well-formed tree). We solve the following Integer Linear Program (ILP)\footnote{There are a number of ways to reconcile the span conflicts, including an adaptation of the standard dynamic programming chart parsing algorithm to work with spans of an unbinarized tree. However it turns out that the classification model rarely produces span conflicts, so all methods we tried performed equivalently well.} to find the highest scoring parse that admits a well-formed tree:

\begin{equation*} \label{eq:1}
\max_\delta \sum_{(i, j) \in \mathsf{spans}(x)} v^+_{(i, j)} \delta_{(i, j)} + v^-_{(i, j)} (1 - \delta_{(i, j)})
\end{equation*}
subject to:
\begin{eqnarray*}
i < k < j < m &\implies& \delta_{(i, j)} + \delta_{(k, m)} \leq 1 \\
(i,j) \in \mathsf{spans}(x) &\implies& \delta_{(i, j)} \in \{0, 1\} 
\end{eqnarray*}

\noindent where:

\begin{eqnarray*}
v^+_{(i, j)} &=& \max_{l \text{ s.t. } l \neq \varnothing} \sigma(l \mid x, (i, j))\\
v^-_{(i, j)} &=& \sigma(\varnothing \mid x, (i, j))
\end{eqnarray*}

\subsection{Classification Model}

% \begin{figure*}[tb]
% \centering
%   \includegraphics{lstm-sketch3.pdf}
%   \caption{A sketch of how a span is encoded by the classification model.}
%   \label{fig:model-sketch}
% \end{figure*}

For our span classification model $\sigma(l \mid x, s)$, we use the model from \citep{DBLP:journals/corr/SternAK17}, which leverages a method for encoding spans from \citep{wang2016graph,Cross2016SpanBasedCP}. First, it creates a sentence encoding by running a two-layer bidirectional LSTM over the sentence to obtain forward and backward encodings for each position $i$, denoted by $\bm{f}_i$ and $\bm{b}_i$ respectively. Then, spans are encoded by the difference in LSTM states immediately before and after the span; that is, span $(i, j)$ is encoded as the concatenation of the vector differences $\bm{f}_j - \bm{f}_{i-1}$ and $\bm{b}_i - \bm{b}_{j+1}$. A one-layer feedforward network maps each span representation to a distribution over labels. 

\subsubsection*{Classification Model Parameters and Initializations}
We preserve the settings used in \citet{DBLP:journals/corr/SternAK17} where possible. As a result, the size of the hidden dimensions of the LSTM and the feedforward network is $250$. The dropout ratio for the LSTM is set to $0.4~$. Unlike the model it is based on, our model uses word embeddings of length $1124$. These result from concatenating a $100$ dimension learned word embedding, with a $1024$ dimension learned linear combination of the internal states of a bidirectional language model run on the input sentence as described in \citet{2018arXiv180205365P}. We refer to them below as ELMo (Embeddings
from Language Models). For the learned embeddings, words with $n$ occurrences in the training data are replaced by $\langle \text{UNK} \rangle$ with probability $\frac{1 + \frac{n}{10}}{1 + n}$. This does not affect the ELMo component of the word embeddings. As a result, even common words are replaced with probability at least $\frac{1}{10}$, making the model rely on the ELMo embeddings instead of the learned embeddings. To make the model self-contained, it does not take part-of-speech tags as input. Using a linear layer over the last hidden layer of the classification model, part-of-speech tags are predicted for spans containing single words.

\section{Analysis of RSP}

\subsection{Performance on Newswire}

% We begin our analysis with the feature ablation in Table~\ref{fig:ablation}. We discover that with ELMo embeddings, there is no longer any benefit to pre-tagging sentences with a POS tagger (as was necessary for MSP). \mpcomment{I don't see how this is shown i Table 1? Where is the ablation with ELMo but without POS tagger?} Therefore, in all following experiments, RSP will refer to a Reconciled Span Parser without pretagging.

On \textsc{WSJTest}\footnote{For all our experiments on the WSJ component of the Penn Treebank \citep{marcus1993building}, we use the standard split which is sections 2-21 for training, henceforth \textsc{WSJTrain}, section 22 for development, henceforth \textsc{WSJDev}, and 23 for testing, henceforth \textsc{WSJTest}. }, RSP outperforms (see Table~\ref{ptb-single-model}) all previous single models trained on \textsc{WSJTrain} by a significant margin, raising the state-of-the-art result from 92.6\% to 94.3\%. Additionally, our predicted part-of-speech tags achieve 97.72\%\footnote{The split we used is not standard for part-of-speech tagging. As a result, we do not compare to part-of-speech taggers.} accuracy on \textsc{WSJTest}.

\begin{table}[t!]
\begin{center}
\begin{tabular}{c|c c c}
\hline 
\bf Parser & \bf Rec & \bf Prec & $\mathbf{F_1}$ \\ 
\hline
%\cite{DBLP:journals/corr/VinyalsKKPSH14} & - & - & 88.3 \\
RNNG \citep{DBLP:journals/corr/DyerKBS16} & - & - & 91.7 \\
MSP \citep{DBLP:journals/corr/SternAK17} & 90.6 & 93.0 & 91.8 \\
\citep{DBLP:conf/emnlp/SternFK17} & 92.6 & 92.6 & 92.6 \\
\hline
RSP & 93.8 & 94.8 & 94.3 \\
\hline
\end{tabular}
\end{center}
\caption{\label{ptb-single-model} Parsing performance on \textsc{WSJTest}, along with the results of other recent single-model parsers trained without external parse data.}
\end{table}

\begin{table}[t!]
\begin{center}
\begin{tabular}{c|c c c}
\hline 
& \bf Recall & \bf Precision & \bf F1 \\ 
\hline
all features & 94.20 & 94.77 & 94.48 \\
--ELMo & 91.63 & 93.05 & 92.34 \\
\hline
\end{tabular}
\end{center}
\caption{\label{fig:ablation} Feature ablation on \textsc{WSJDev}.}
\end{table}

% Converting these constituency parses into dependency parses\footnote{To do so, we use the Stanford NLP toolkit (\textbf{CITE}).} gets us \textbf{TODO} UAS and \textbf{TODO} LAS, again raising the state-of-the-art from 95.8\% UAS and 94.6\% LAS \cite{DBLP:conf/eacl/SmithDBNKK17}.
%Note that this is not a single model since it uses a reranker.
% Moreover, Table~\ref{ptb-pos-acc} shows that the POS tags predicted by RSP on \textsc{WSJTest} are competitive with state-of-the-art taggers.
% part-of-speech tag split is 0-18 train, 19-21 dev, and 22-24 test

% \begin{table}[t!]
% \begin{center}
% \begin{tabular}{c|c}
% \hline 
% \bf Model & \bf Acc. \\ 
% \hline
% %https://openreview.net/pdf?id=SJTCsqMUf
% Collobert et al. ~\shortcite{DBLP:journals/jmlr/CollobertWBKKK11} & 97.27 \\
% Ma \& Hovy et al. ~\shortcite{DBLP:conf/acl/MaH16} & 97.55 \\
% Ling et al. ~\shortcite{DBLP:conf/emnlp/LingDBTFAML15} & 97.78 \\
% \hline
% Our Model & 97.72 \\
% \hline
% \end{tabular}
% \end{center}
% \caption{\label{ptb-pos-acc} Part-of-Speech Tagging Accuracy on Penn Treebank. }
% \end{table}

%We also confirm that the use of optimal decoding in RSP does not result in a significant slowdown compared to MSP. Parsing \textsc{WSJTest} with RSP\footnote{\textbf{Describe computer architecture.}} takes \textbf{TODO} seconds, compared to \textbf{TODO} seconds for MSP.

\subsection{Beyond Newswire}

\subsubsection*{The Brown Corpus}

The Brown corpus \citep{marcus1993building} is a standard benchmark used to assess WSJ-trained parsers outside of the newswire domain. When \citep{Kummerfeld2012ParserSA} parsed the various Brown verticals with the (then state-of-the-art) Charniak parser \citep{Charniak2000AMP,charniak2005coarse, McClosky2006EffectiveSF}, it achieved $F_1$ scores between 83\% and 86\%, even though its $F_1$ score on \textsc{WSJTest} was 92.1\%.

In Table~\ref{brown-corpus}, we discover that RSP does not suffer nearly as much degradation, with an average $F_1$-score of $90.3\%$. To determine whether this increased portability is because of the parser architecture or the use of ELMo vectors, we also run MSP on the Brown verticals. We used the Stanford tagger\footnote{We used the english-left3words-distsim.tagger model from the 2017-06-09 release of the Stanford POS tagger since it achieved the best accuracy on the Brown corpus.} \citep{DBLP:conf/naacl/ToutanovaKMS03} to tag \textsc{WSJTrain} and the Brown verticals so that MSP could be given these at train and test time. We learned that most of the improvement can be attributed to the ELMo word representations. In fact, even if we use MSP with gold POS tags, the average performance is $3.4\%$ below RSP.

\begin{table*}
\centering
\renewcommand{\arraystretch}{1.1}
\begin{tabular}{c c c c c}
\toprule
\textbf{Section} & \multicolumn{4}{c}{$\mathbf{F_1}$} \\
& \textbf{RSP} & \textbf{MSP + Stanford POS tags} & \textbf{MSP + gold POS tags} & \textbf{Charniak}\\
\midrule
F (popular) & 91.42 & 87.01 & 87.84 & 85.91 \\
G (biographies) & 90.04 & 86.14 & 86.91 & 84.56\\
K (general) & 90.08 & 85.53 & 86.46 & 84.09 \\
L (mystery) & 89.65 & 85.61 & 86.47 & 83.95 \\
M (science) & 90.52 & 86.91 & 87.52 & 84.65 \\
N (adventure) & 91.00 & 86.53 & 87.53 & 85.2 \\
P (romance) & 89.76 & 85.77 & 86.59 & 84.09 \\
R (humor) & 89.54 & 84.98 & 85.69 & 83.60 \\
\midrule
average & 90.25 & 86.06 & 86.88 & 84.51 \\
\bottomrule
\end{tabular}
\caption{\label{brown-corpus} Parsing performance on Brown verticals. MSP refers to the Minimal Span Parser \citep{DBLP:journals/corr/SternAK17}.  Charniak refers to the Charniak parser with reranking and self-training \citep{Charniak2000AMP,charniak2005coarse, McClosky2006EffectiveSF}. MSP + Stanford POS tags refers to MSP trained and tested using part-of-speech tags predicted by the Stanford tagger \citep{DBLP:conf/naacl/ToutanovaKMS03}.}
\end{table*}

\subsubsection*{Question Bank and Genia}

\begin{table}[t!]
\begin{center}
\begin{tabular}{c c|c c c}
\hline 
\multicolumn{2}{c|}{\textbf{Training Data}} & \bf Rec. & \bf Prec. & $\mathbf{F_1}$ \\ 
\textsc{WSJ} & \textsc{QBank} & & & \\
\hline
\hline
40k & 0 & 91.07 & 88.77 & 89.91 \\
0 & 2k & 94.44 & 96.23 & 95.32 \\
40k & 2k & 95.84 & 97.02 & 96.43 \\
\hline
40k & 50 & 93.85 & 95.91 & 94.87 \\
40k & 100 & 95.08 & 96.06 & 95.57 \\
40k & 400 & 94.94 & 97.05 & 95.99 \\
\hline
\end{tabular}
\end{center}
\caption{\label{qbanksubsets} Performance of RSP on \textsc{QBankDev}. }
\end{table}

\begin{table}[t!]
\begin{center}
\begin{tabular}{c c|c c c}
\hline 
\multicolumn{2}{c|}{\textbf{Training Data}} & \textbf{Rec} & \textbf{Prec} & $\mathbf{F_1}$ \\ 
\textsc{WSJ} & \textsc{Genia} & \\
\hline
\hline
40k & 0 & 72.51 & 88.84 & 79.85 \\
0k & 14k & 88.04 & 92.30 & 90.12 \\
40k & 14k & 88.24 & 92.33 & 90.24 \\
\hline
40k & 50 & 82.30 & 90.55 & 86.23 \\
40k & 100 & 83.94 & 89.97 & 86.85 \\
40k & 500 & 85.52 & 91.01 & 88.18 \\
\hline
\end{tabular}
\end{center}
\caption{\label{geniasubsets} Performance of RSP on \textsc{GeniaDev}. }
\end{table}

Despite being a standard benchmark for parsing domain adaptation, the Brown corpus has considerable commonality with newswire text. It is primarily composed of well-formed sentences with similar syntactic phenomena. Perhaps the main challenge with the Brown corpus is a difference in vocabulary, rather than a difference in syntax, which may explain the success of RSP, which leverages contextualized embeddings learned from a large corpus.

If we try to run RSP on a more syntactically divergent corpus like QuestionBank\footnote{For all our experiments on QuestionBank, we use the following split: sentences 1-1000 and 2001-3000 for training, henceforth \textsc{QBankTrain}, 1001-1500 and 3001-3500 for development, henceforth \textsc{QBankDev}, and 1501-2000 and 2501-4000 for testing, henceforth \textsc{QBankTest}. This split is described at https://nlp.stanford.edu/data/QuestionBank-Stanford.shtml.} \citep{judge2006questionbank}, we find much more performance degradation. This is unsurprising, since \textsc{WSJTrain} does not contain many examples of question syntax. But how many examples do we need, to get good performance?

For the experiments summarized in table~\ref{qbanksubsets} and table~\ref{geniasubsets} involving 40k sentences from \textsc{WSJTrain}, we started with RSP trained on \textsc{WSJTrain}, and fine-tuned it on minibatches containing an equal number of target domain and \textsc{WSJTrain} sentences. 

Surprisingly, with only 50 annotated questions (see Table~\ref{qbanksubsets}), performance on \textsc{QBankDev} jumps 5 points, from 89.9\% to 94.9\%. This is only 1.5\% below training with all of \textsc{WsjTrain} and \textsc{QBankTrain}. The resulting system improves slightly on \textsc{WsjTest} getting 94.38\%.

On the more difficult \textsc{Genia} corpus of biomedical abstracts \citep{tateisi2005syntax}, we see a similar, if somewhat less dramatic, trend. See Table~\ref{geniasubsets}. With 50 annotated sentences, performance on \textsc{GeniaDev} jumps from 79.5\% to 86.2\%, outperforming all but one parser from David McClosky's thesis \citep{mcclosky2010any} -- the one that trains on all 14k sentences from \textsc{GeniaTrain} and self-trains using 270k sentences from PubMed. That parser achieves 87.6\%, which we outperform with just 500 sentences from \textsc{GeniaTrain}.

These results suggest that it is currently feasible to extend a parser to a syntactically distant domain (for which no gold parses exist) with a couple hours of effort. We explore this possibility in the next section.

\begin{figure}
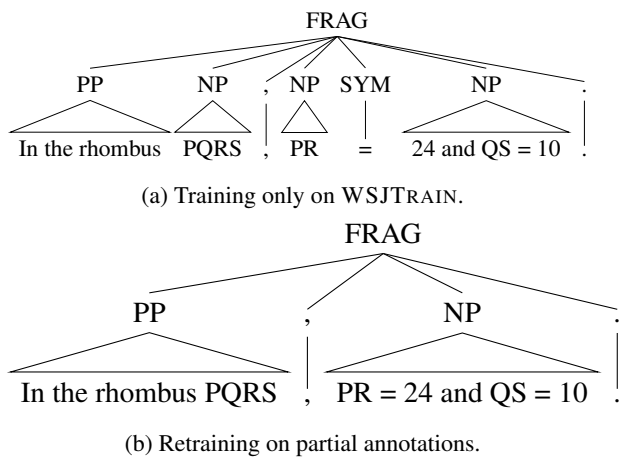

\centering
\begin{subfigure}{\linewidth}
  \centering
  \resizebox{\linewidth}{!}{%
  \Tree [.FRAG  [.PP \edge[roof]; {In the rhombus} ]
			  [.NP \edge[roof]; {PQRS} ]
              [., , ]
              [.NP \edge[roof]; {PR} ]
              [.SYM = ]
              [.NP \edge[roof]; {24 and QS = 10} ]
              [.. . ]]
  }
  \caption{Training only on \textsc{WSJTrain}.}
  \label{fig:pre-training-geo}
\end{subfigure} 
\begin{subfigure}{\linewidth}
  \centering
  \Tree [.FRAG [.PP \edge[roof]; {In the rhombus PQRS} ]
			 [., , ]
             [.NP \edge[roof]; {PR = 24 and QS = 10} ]
             [.. . ]]
  \caption{Retraining on partial annotations.}
  \label{fig:post-training-geo}
\end{subfigure}
\caption{\label{fig:geotrained} The top-level split for the development sentence ``In the rhombus PQRS, PR = 24 and QS = 10.'' before and after retraining RSP on 63 partially annotated geometry statements.}
\end{figure}

\section{Rapid Parser Extension}

To create a parser for their geometry question answering system, \cite{Seo2015SolvingGP} did the following:

\begin{itemize}
	\item Designed regular expressions to identify mathematical expressions.
	\item Replaced the identified expressions with dummy words.
	\item Parsed the resulting sentences.
	\item Substituted the regex-analyzed expressions for the dummy words in the parses. 
\end{itemize}

\noindent It is clear why this was necessary. Figure~\ref{fig:geotrained} (top) shows how RSP (trained only on \textsc{WSJTrain}) parses the sentence ``In the rhombus PQRS, PR = 24 and QS = 10." The result is completely wrong, and useless to a downstream application.

\begin{table*}[t]
\centering
\renewcommand{\arraystretch}{1.1}
\begin{tabular}{c|c c|c}
\hline
Training Data & \multicolumn{2}{c|}{\textsc{GeoDev}} & \textsc{WsjTest} \\
& correct constituents \% & error-free \% & $F_1$ \\
\hline
\textsc{WsjTrain} & 71.9 & 45.2 & 94.28  \\
\textsc{WsjTrain} + \textsc{GeoTrain} & 87.0 & 72.6 & 94.30  \\
\hline
\end{tabular}
\caption{\label{tab:georesults} RSP performance on \textsc{GeoDev}.}
\end{table*}

\begin{table*}[t]
\centering
\renewcommand{\arraystretch}{1.1}
\begin{tabular}{c|c c|c}
\hline
Training Data & \multicolumn{2}{c|}{\textsc{BioChemDev}} & \textsc{WsjTest} \\
& correct constituents \% & error-free \% & $F_1$ \\
\hline
\textsc{WsjTrain} & 70.1 & 27.0 & 94.28  \\
\textsc{WsjTrain} + \textsc{BioChemTrain} & 79.5 & 46.7 & 94.23  \\
\hline
\end{tabular}
\caption{\label{tab:biochemresults} RSP performance on \textsc{BioChemDev}.}
\end{table*}

Still, beyond just the inconvenience of building additional infrastructure, there are downsides to the ``regex-and-replace" strategy:

\begin{enumerate}
	\item \textbf{It assumes that each expression always maps to the same constituent label.} Consider ``$2x=3y$". This is a verb phrase in the sentence ``In the above figure, x is prime and $2x=3y$." However, it is a noun phrase in the sentence ``The equation $2x=3y$ has 2 solutions." If we replace both instances with the same dummy word, the parser will almost certainly become confused in one of the two instances.
	\item \textbf{It assumes that each expression is always a constituent.} Suppose that we replace the expression ``$AB < 30$" with a dummy word. This means we cannot properly parse a sentence like ``When angle $AB < 30$, the lines are parallel," because the constituent ``angle $AB$" no longer exists in the resulting sentence.
	\item \textbf{It does not handle other syntactic variation.} As we will see in the next section, the geometry domain has a propensity for using right-attaching participial adjective phrases, like ``labeled x'' in the phrase ``the segment labeled x." Encouraging a parser to recognize this syntactic construct is out-of-scope for the ``regex-and-replace" strategy.
\end{enumerate}

\noindent Instead, we propose directly extending the parser by providing a few domain-specific examples like those in Figure~\ref{fig:markups}. Because RSP's model directly predicts span constituency, we can simply mark up a sentence with the ``tricky" domain-specific constituents that the model will not already have learned from \textsc{WSJTrain}. For instance, we mark up \textsc{Noun}-\textsc{Label} constructs like ``chord BD", and equations like ``AD = 4".

From these marked-up sentences, we can extract training instances declaring the constituency of certain spans (like ``to chord BD" in the third example) and the implied non-constituency of certain spans (like ``perpendicular to chord" in the third example). We also allow annotators to explicitly declare the non-constituency of a span via an alternative markup (not shown).

We do not require annotators to provide span labels (although they can if desired). If a training instance merely declares a span to be a constituent (but does not provide a particular label), then the loss function only records loss when that span is classified as a non-constituent (i.e. any label is ok).

% Also, because RSP does not require binarization of the parse trees for decoding (unlike MSP), an annotator does not need to understand the unintuitive concept of virtual spans introduced by binarization. Rather, in order to annotate effectively, an annotator only needs some awareness of the coarse syntactic chunks of a sentence.

\section{Experiments}
\label{sec:exp}

\subsection{Geometry Questions}

We took the publicly available training data from \cite{Seo2015SolvingGP}, split the data into sentences, and then annotated each sentence as in Figure~\ref{fig:markups}. Next, we randomly split these sentences into \textsc{GeoTrain} and \textsc{GeoDev}\footnote{\textsc{GeoTrain} and \textsc{GeoDev} are available at https://github.com/vidurj/parser-adaptation/tree/master/data.}. After removing duplicate sentences spanning both sets, we ended up with 63 annotated sentences in \textsc{GeoTrain} and 62 in \textsc{GeoDev}. In \textsc{GeoTrain}, we made an average of 2.8 constituent declarations and 0.3 (explicit) non-constituent declarations per sentence.

After preparing the data, we started with RSP trained on \textsc{WsjTrain}, and fine-tuned it on minibatches containing 50 randomly selected \textsc{WsjTrain} sentences, plus all of \textsc{GeoTrain}. The results are in table~\ref{tab:georesults}. After fine-tuning, the model gets 87\% of the 185 annotations on \textsc{GeoDev} correct, compared with 71.9\% before fine-tuning\footnote{This improvement has a p-value of $10^{-4}$ under the one-sided, two-sample difference between proportions test.}. Moreover, the fraction of sentences with no errors increases from 45.2\% to 72.6\%.  With only a few dozen partially-annotated training examples, not only do we see a large increase in domain performance, but there is also no degradation in the parser's performance on newswire. Some \textsc{GeoDev} parses have enormous qualitative differences, like the example shown in Figure~\ref{fig:geotrained}.

\begin{figure}[tb]
\begin{mdframed}[roundcorner=10pt]
\begin{itemize}
\item \textbf{Given [ a circle with [ the tangent shown ] ] .}
\item \textbf{Find the hypotenuse of [ the triangle labeled t ] .}
\item \textbf{Examine [ the following diagram with [ the square highlighted ] ] .}
\end{itemize}
\end{mdframed}
\caption{\label{fig:iter2} Three partial annotations targeting right-attaching participial adjectives.}
\end{figure}

For the \textsc{GeoDev} sentences on which we get errors after retraining, the errors fall predominantly into three categories. First, approximately 44\% have some mishandled math syntax, like failing to recognize ``dimensions 16 by 8" as a constituent, or providing a flat structuring of the equation ``BAC = 1/4 * ACB" (instead of recognizing ``1/4 * ACB" as a subconstituent). Second, approximately 19\% have PP-attachment errors. Third, another 19\% fail to correctly analyze right-attaching participial adjectives like ``labeled x" in the noun phrase ``the segment labeled x" or ``indicated" in the noun phrase ``the center indicated." This phenomenon is unusually frequent in geometry but was insufficiently marked-up in our training examples. For instance, while we have a training instance ``Find [ the measure of [ the angle designated by x ] ]," it does not explicitly highlight the constituency of ``designated by x".  This suggests that in practice, this domain adaptation method could benefit from an iterative cycle in which a user assesses the parser's errors on their target domain, creates some partial annotations that address these issues, retrains the parser, and then repeats the process until satisfied. As a proof-of-concept, we invented 3 additional sentences with right-attaching participial adjectives (shown in Figure~\ref{fig:iter2}), added them to \textsc{GeoTrain}, and then retrained. Indeed, the handling of participial adjectives in \textsc{GeoDev} improved, increasing the overall percentage of correctly identified constituents to 88.6\% and the percentage of error-free sentences to 75.8\%.

\subsection{Biomedicine and Chemistry}

We ran a similar experiment using biomedical and chemistry text, taken from the unannotated data provided by \cite{nivre2007conll}. We partially annotated 134 sentences and randomly split them into \textsc{BioChemTrain} (72 sentences) and \textsc{BioChemDev} (62 sentences)\footnote{\textsc{BioChemTrain} and \textsc{BioChemDev} are available at https://github.com/vidurj/parser-adaptation/tree/master/data.}. In \textsc{BioChemTrain}, we made an average of 4.2 constituent declarations per sentence. We made no non-constituent declarations.

Again, we started with RSP trained on \textsc{WsjTrain}, and fine-tuned it on minibatches containing annotations from 50 randomly selected \textsc{WsjTrain} sentences, plus all of \textsc{BioChemTrain}. Table~\ref{tab:biochemresults} shows the improvement in the percentage of correctly-identified annotated constituents and the percentage of test sentences for which the parse agrees with every annotation. As with the geometry domain, we get significant improvements using only dozens of partially annotated training sentences.

\section{Related Work}

The two major themes of this paper, domain adaptation and learning from partial annotation, each have a long tradition in natural language processing.

\subsection{Domain Adaptation}

Domain adaptation has been recognized as a major NLP problem for over a decade \cite{BenDavid2006AnalysisOR,Blitzer2006DomainAW,Daum2007FrustratinglyED,Finkel2009HierarchicalBD}. In particular, domain adaptation for parsers \cite{plank2011domain,Ma2013DependencyPA} has received considerable attention. Much of this work \cite{McClosky2006RerankingAS,Reichart2007SelfTrainingFE,Sagae2007DependencyPA,Kawahara2008LearningRO,McClosky2010AutomaticDA,Sagae2010SelfTrainingWR,Baucom2013DomainAF,Yu2015DomainAF} has focused on how to best use co-training \cite{blum1998combining} or self-training to augment a small domain corpus, or how to best combine models to perform well on a particular domain.

In this work, we focus on the direct impact that just a few dozen partially annotated out-of-domain examples can have, when using a particular neural model with contextualized word representations. Co-training, self-training, and model combination are orthogonal to our approach. Our work is a spiritual successor to \cite{garrette2013learning}, which shows how to train a part-of-speech tagger with a minimal amount of annotation effort.

\subsection{Learning from Partial Annotation}

Most literature on training parsers from partial annotations \cite{Sassano2010UsingSC,Spreyer2010TrainingPO,Flannery2011TrainingDP,Flannery2015CombiningAL,Mielens2015ParseIF} focuses on dependency parsing. \cite{Li2016TrainingDP} provides a good overview. Here we highlight three important high-level strategies.

The first is ``complete-then-train" \cite{Mirroshandel2011ActiveLF,Majidi2013CommitteeBasedAL}, which ``completes" every partially annotated dependency parse by finding the most likely parse (according to an already trained parser model) that respects the constraints of the partial annotations. These ``completed" parses are then used to train a new parser.

The second strategy \cite{Nivre2014ConstrainedAD,Li2016TrainingDP} is similar to ``complete-then-train," but integrates parse completion into the training process. At each iteration, new ``complete" parses are created using the parser model from the most recent training iteration.

The third strategy \cite{Li2014SoftCS,Li2016TrainingDP} transforms each partial annotation into a forest of parses that encodes all fully-specified parses permitted by the partial annotation. Then, the training objective is modified to support optimization over these forests.

Our work differs from these in two respects. First, since we are training a constituency parser, our partial annotations are constituent bracketings rather than dependency arcs. Second, and more importantly, we can use the partial annotations for training without modifying either the training algorithm or the training data. 

While the bulk of the literature on training from partial annotations focuses on dependency parsing, the earliest papers \cite{Pereira1992InsideOutsideRF,Hwa1999SupervisedGI} focus on constituency parsing. These leverage an adapted version of the inside-outside algorithm for estimating the parameters of a probabilistic context-free grammar (PCFG). Our work is not tied to PCFG parsing, nor does it require a specialized training algorithm when going from full annotations to partial annotations.

\section{Conclusion}

Recent developments in neural natural language processing have made it very easy to build custom parsers. Not only do contextualized word representations help parsers learn the syntax of new domains with very few examples, but they also work extremely well with parsing models that correspond directly with a granular and intuitive annotation task (like identifying whether a span is a constituent). This allows you to train with either full or partial annotations without any change to the training process.

This work provides a convenient path forward for the researcher who requires a parser for their domain, but laments that ``parsers don't work outside of newswire." With a couple hours of effort (and a layman's understanding of syntactic building blocks), they can get significant performance improvements. We envision an iterative use case in which a user assesses a parser's errors on their target domain, creates some partial annotations to teach the parser how to fix these errors, then retrains the parser, repeating the process until they are satisfied.

\bibliographystyle{acl_natbib}
\bibliography{acl2018}

\end{document}